\pdfoutput=1

\documentclass[10pt]{article} % For LaTeX2e

\usepackage[preprint]{rlj}

\usepackage{amssymb}            % Defines common symbols like \mathbb R
\usepackage{mathtools}          % Extends amsmath, providing common math tools
\usepackage{mathrsfs}           % Enables \mathscr, which can work in cases that \mathcal does not
\usepackage{graphicx}           % For including images
\usepackage{subcaption}         % Allows for the use of subfigures and subcaptions
\usepackage[space]{grffile}     % For spaces in image names
\usepackage{url}                % For displaying URLs
\usepackage{cleveref}           % For \Cref and enhanced cross-referencing
\usepackage{booktabs}
\usepackage{pdflscape}
\usepackage{rotating}
\usepackage{multirow}
\usepackage{wrapfig}
\usepackage{algpseudocode}
\newcommand{\btob}{{\textit{B2B}}}
\newcommand{\buildingtobuilding}{{Building2Building}}

\title{Building2Building: A Large Scale Benchmark for Generalizable Real-World Reinforcement Learning}

\setrunningtitle{Building2Building: A Benchmark for Generalizable Reinforcement Learning at Scale}

\author{Vincent Taboga\textsuperscript{1,2}, Justin Veilleux \textsuperscript{1,2}, Doseok Jang\textsuperscript{1,2}, Anushree Rankawat\textsuperscript{2}, Pierre-Luc Bacon\textsuperscript{1,2}}

\emails{vincent.taboga@polymtl.ca}

\affiliations{
$^{1}$\textbf{Mila - Quebec AI Institute}\\
$^{2}$\textbf{Université de Montréal - Département d'Informatique et de Recherche Opérationnelle}\\
}

\keywords{Reinforcement Learning; Generalization; Multi-Task; Transfer Learning; Benchmark; Energy; HVAC}

\begin{document}

\maketitle

\begin{abstract}
Reinforcement learning (RL) has achieved strong results in control, yet learned policies remain brittle to changes in dynamics, action spaces, observation spaces, or goals, a critical limitation for real-world deployment. Existing benchmarks offer limited diversity and complexity, making it difficult to rigorously study transfer, multi-task learning, and meta-learning in RL. We introduce \buildingtobuilding{}\footnote{Code available at \url{https://github.com/vtaboga/building2building}} (\btob{}), a large-scale suite of realistic Heating, Ventilation, and Air Conditioning (HVAC) control environments built on EnergyPlus, a state-of-the-art building simulator. \btob{} is fully compatible with the Gymnasium interface and features a parametric building generator, enabling the systematic generation of diverse building configurations with heterogeneous observation and action spaces. Based on this suite, we define benchmark tasks targeting key open challenges in RL, including goal adaptation, dynamics adaptation, action-space shifts, and cross-domain transfer. By providing a large-scale, diverse, and physically grounded testbed with standardized evaluation protocols, \btob{} enables systematic investigation of generalization and transfer in continuous control. Beyond advancing research on generalization in RL, this new benchmark also carries significant societal implications by enabling improved HVAC control at scale, one of the most energy-intensive systems in buildings.
\end{abstract}

\section{Introduction}
\label{sec:introduction}

Reinforcement Learning (RL) has achieved strong performance across a wide range of control problems. In most cases, however, these results are obtained by training a policy on a single target task. While effective in controlled settings, this paradigm limits the ability of RL systems to operate in new environments or adapt to related tasks. Adaptation is a key feature for real-world deployments, and enabling RL agents to generalize and adapt across various environments remains a major open challenge.

Studying generalization in RL requires simulated environments that expose agents to many tasks with varying dynamics, observation spaces, action spaces, and objectives. Existing benchmarks such as Meta-World+ \citep{mclean2025} and RLBench \citep{stephen2020} provide useful testbeds, but remain largely confined to the robotics domain. Moreover, the variations they introduce are often limited to changes in dynamics parameters (e.g., friction) or task goals such as target positions for navigation or object placement. As a result, these benchmarks offer limited diversity in environment structure and rarely include heterogeneous observation and action spaces. Because of this limitation, many works studying cross-domain adaptation or transfer in RL rely on custom experimental setups. These environments are often designed for a specific method and are difficult to reproduce, which makes comparisons across approaches challenging. Consequently, progress toward generalizable RL agents remains fragmented.

In this work, we propose a new domain for studying generalization in RL: heating, ventilation, and air-conditioning (HVAC) control in buildings. Buildings exhibit large diversity in layouts, materials, climate exposure, HVAC configurations and objectives balancing energy savings and comfort. At the same time, they share common physical principles based on thermodynamics and energy conservation. This combination of diverse dynamics, heterogeneous control interfaces, and shared physical principles makes HVAC control an ideal domain for studying generalization.

We introduce \buildingtobuilding{} (\btob{}), a large-scale suite of RL environments built on EnergyPlus, a state-of-the-art building energy simulator \citep{energyplus}. EnergyPlus provides high-fidelity physical modeling of building thermal behavior. The simulator's accuracy is highlighted by works such as \citet{zhang2019}, which showed that RL policies trained in calibrated EnergyPlus simulations can be deployed directly in real buildings. 

Our environment suite includes a building generator that produces a virtually infinite number of environments derived from base archetypes across several geographic locations in North America. These environments expose low-level HVAC actuators such as fans and dampers, creating a wide variety of challenging control problems with heterogeneous observation and action spaces, and objectives reflecting different comfort-energy trade-offs. These models are automatically converted into Gymnasium-compatible RL environments through a dedicated processing pipeline. 

Using this generator, we construct a dataset of 6000 environments spanning multiple building types, climates, and HVAC system configurations. We further propose well-defined and reproducible benchmark problems that isolate different sources of variation in RL: goal adaptation, dynamics variation, action-space shifts, and cross-domain generalization.

By providing a large and diverse collection of realistic control environments, \btob{} enables systematic and large-scale evaluation of generalization methods in RL. At the same time, it supports research on intelligent HVAC control and helps bridge the gap between fundamental RL research and real-world deployment on energy management systems. HVAC control represents an important societal challenge. Buildings account for roughly one third of global energy consumption, and HVAC systems alone are responsible for nearly half of this usage. Improving HVAC control can therefore reduce energy consumption, lower operational costs, and contribute directly to climate mitigation \citep{iea_outlook2022}.

\section{Related Work}

\subsection{Environments and Benchmarks for Generalization in RL}
Existing benchmarks for studying generalization in RL are summarized in \Cref{tab:rl_benchmarks}, and the different approaches to the generalization problem in RL are further discussed in \Cref{apdx:prior_work_generalization_rl}. Most of the benchmarks focus on either video games or robotics tasks such as locomotion and manipulation. As a result, they cover only a limited set of application domains. These problems have been extensively studied over the past decade. While they have led to significant algorithmic progress, it remains unclear whether these improvements generalize beyond the specific domains on which they are evaluated. As highlighted by \citet{liao2021}, repeatedly evaluating algorithms on the same benchmarks may lead to overfitting to the benchmark itself rather than genuine progress in general RL capabilities. This motivates the need for new environments covering different applications and offering large diversity of tasks.

\subsection{HVAC control}
Much of the existing literature on RL for HVAC follows a familiar pattern: select a single building, connect it to a simulator, and demonstrate that an agent can reduce energy consumption while maintaining thermal comfort. Surveys show that this type of study has become common practice \citep{sierla2022,slsayed2024}. Optimal control of HVAC systems benefits from a rich literature, and \citet{Xu2025} propose to leverage expert knowledge to improve the sample efficiency of online methods. However, scaling online approaches beyond individual case studies remains difficult. Each new building requires its own digital twin, with sensors, actuators, and dynamics configured for simulation, and policies trained in one building rarely transfer to others.

Existing environments, summarized in \Cref{tab:hvac_envs}, \Cref{apdx:environments_details} reflect this limitation. Platforms such as Sinergym \citep{campoy2025}, Energym \citep{scharnhorst2021}, BOPTEST \citep{blum2021}, and CityLearn \citep{nweye2024} have accelerated research but collectively provide only a few dozen building models. Even within this limited set, policies trained in one environment often fail to generalize to others \citep{manjavacas2024}. Transfer-learning approaches, such as fine-tuning across buildings \citep{kadamala2024,xu2020,coraci2024} or continual learning with hypernetworks \citep{bekal2025}, provide incremental improvements but remain constrained by the lack of large-scale benchmarks. Field studies report similar challenges: a review of more than one hundred deployments finds inconsistent savings and limited reporting on integration costs \citep{habbazi2025}.

In contrast, \btob{} enables large-scale studies of RL for HVAC control. This capability enables a new training paradigm for HVAC control. Rather than designing a highly detailed digital twin for each individual building, policies can be trained across a distribution of buildings that includes the target building. Training across many environments reduces the need for highly accurate per-building models and allows policies to reuse experience across environments, reducing the need to train a new controller from scratch for every building.

\begin{table}
\centering
\footnotesize
\caption{Main benchmarks used to study generalization in reinforcement learning.}
\begin{tabular}{l l l}
\hline
\textbf{Benchmark} & \textbf{Tasks} & \textbf{Domain} \\
\hline
MineRL \citep{guss2019} & open / task generators & Video game (Minecraft) \\
Procgen \citep{cobbe2020} & 16 task families & Procedural video games \\
RLBench \citep{stephen2020} & 100 tasks & Robotics manipulation \\
Meta-World \citep{mclean2025} & 50 tasks & Robotics manipulation \\
DM Control \citep{tunyasuvunakool2022} & $\sim$20 tasks & Robotics locomotion / control \\
ManiSkill2 \citep{gu2023} & 20 task families & Robotics manipulation \\
Robosuite \citep{yuke2020} & 9 tasks & Robotics manipulation \\
MMBench \citep{hansen2026} & 200 tasks & Games and robotics \\
Habitat \citep{szot2022} & open / configurable & Embodied navigation \\
\hline
\end{tabular}
\label{tab:rl_benchmarks}
\end{table}

\section{The Building2Building Suite}
\label{sec:B2B_suite}

\subsection{Environment Generation}

To access a diverse set of environments representative of real buildings, we rely on two complementary sources of building models that produce a large set of heterogeneous environments, summarized in \Cref{tab:base_environments}\footnote{Observation dimensions are given for constant-setpoint tasks; occupancy-based and
random-setpoint tasks add one occupancy and one target-temperature signal per controlled zone.}. 

\begin{itemize}
    \item A dataset generated by a residential building generator \citep{larochelle_martin2026}. This generator produces single-zone houses representing the statistical characteristics of the Québec residential building stock. 
    \item A parametric building model generator based on the reference commercial buildings from the ASHRAE 90.1 across 16 different locations in North America. The generator follows the methodology used by the Building Technology Assessment Platform \citep{btap2022}, and samples the building size, window-to-wall ratio, insulation levels, and air infiltration. To ensure physical consistency, the generator enforces climate-dependent parameter bounds consistent with ASHRAE standards.
\end{itemize}

\begin{table}
\centering
\small
\caption{Base environments included in the benchmark suite.}
\label{tab:base_environments}
\setlength{\tabcolsep}{4pt}
\begin{tabular}{@{}lcccc@{}}
\toprule
Building type & HVAC type & Controlled zones & Action dim. & Observation dim. \\
\hline
House         & Unitary system               & 1  & 2  & 9--11 \\
Restaurant    & Unitary systems              & 2  & 4  & 10 \\
Warehouse     & Unitary systems \& baseboard & 3  & 5  & 10 \\
Retail Store  & Unitary systems              & 5  & 9  & 12 \\
Small Office  & Unitary systems              & 5  & 10 & 13 \\
Medium Office & Central systems              & 15 & 36 & 25 \\
\bottomrule
\end{tabular}
\end{table}

\subsection{Control Interface}

\btob{} exposes low-level HVAC actuators, and handles natively the following systems:

\begin{enumerate}

\item \textbf{Unitary HVAC systems.}  
These systems condition a single thermal zone using a dedicated unit such as a heat pump or packaged rooftop unit. For these systems, the agent controls the Supply Air Temperature (SAT) and the supply air flow rate.

\item \textbf{Central air loop systems for multi-zone conditioning.}  
These systems use centralized heating and cooling equipment that distributes conditioned air through an air loop serving multiple zones. The agent controls the central SAT, the outdoor-air mixer, the damper opening of each zone, and the zone-level reheat coils through thermostat setpoints.

\item \textbf{Baseboard and heating-only systems.}  
These systems provide heating only and are controlled through the zone thermostat.

\end{enumerate}

At each timestep, the agent has access to the following observation: 
weather data (outdoor air temperature ($^\circ C$) and humidity (\%), calendar information (time of day, day of week, day of year), indoor zones air temperature ($^\circ C$), and HVAC energy consumption per simulation time step ($Wh/m^2/timestep$). Note that the energy consumption is normalized by the floor area of the building to be invariant to building size variations.

\subsection{Control Tasks and Reward Functions}
\label{sec:control_tasks_and_reward_functions}

To study different trade-offs between thermal comfort and energy efficiency, we propose the following reward functions defining a family of control tasks parametrized by energy savings and comfort: 

\begin{equation}
\label{eq:reward_function}
r_t = - \frac{1}{Z \, \tau_T} \sum_{z=1}^Z 
 (T_z(t) - T^{target}_z(t))^2
- \frac{w_E}{\tau_E} \, E_{\text{HVAC}}(t),
\end{equation}

where $T_z$ is the zone temperature, $T_z^{target}$ the zone target temperature, $Z$ denotes the number of zones and $E_{\text{HVAC}}$ the HVAC energy consumption per squared meter. Note that the target temperature $T_z^{target}$ may vary over time depending on occupancy schedules.

Many widely used RL algorithms handle multiple objectives using a weighted sum of rewards \citep{hayes2022practical}. The scale of each term in the sum defines the importance of the objective, and the weights have to be adjusted depending on the desired trade-off. Choosing appropriate weights is difficult and performance is highly sensitive to them \citep{lin2019pareto}. For objectives such as energy consumption for which the scale depends on the HVAC equipment and the building type, the weights are environment dependent, and often tuned manually \citep{togashi2025reward}. Defining the same trade-off with a single set of weights across thousands of environments is extremely challenging, if not impossible, and weights need to be adapted to each environment \citep{vanhasselt2016learning, hessel2018}. To better handle the scale of each reward term, we introduce normalizing constants $\tau_T$ and $\tau_E$ that depend on the building type and the performances of a baseline policy. Using this normalized formulation, each term stays within the same order of magnitude across environments, and a single weight $w_E$ may be used to define a comfort - energy savings trade-off across environments. More details about the reward formulation and normalizing constants are given in \Cref{apdx:reward_normalization}.

\subsection{Structured environment representation}
\label{sec:structured-env}

A major obstacle in developing transfer and multi-task RL algorithms is the lack of standard ways to describe a heterogeneous fleet of different environments. Gymnasium's standard API, which presents an environment through its observation and action spaces, allows writing generic training algorithms and model architectures. However, no equivalent exists for representing families of environments whose action and observation spaces differ, but still form a cohesive family. Without a standard representation for the action and observation spaces, each algorithm for cross-domain transfer defines its own interface, specific to the domain of application. In an effort to facilitate cross-domain transfer research, \btob{} introduces a structured representation that provides a common abstraction for heterogeneous environments.

Rather than reinventing the environment interface, \btob{} extends it: an environment remains a standard Gymnasium environment, and its \emph{morphology} is expressed as a pair of translation procedures between the observation spaces, the action spaces and a decomposed structured form. The decomposition separates each space in two parts: a \emph{common} part shared by the set of environments, and a \emph{morphological graph} carrying information that varies among the environments of the set. Departing from the usual two-space presentation of an environment, three types of data are tracked: observations, actions and \emph{attributes}. \emph{Attributes} are fixed for the lifetime of the environment and describe the environment domain rather than its current state. Attributes play the role of the context in a contextual MDP \citep{hallak2015}: they identify which member of the family the agent is facing.

The vocabulary shared by all members of a family is collected in a \emph{morphological universe}. Formally, a morphological universe $U$ has the following structure:

\begin{enumerate}
    \item A common attribute space $C_c$, a common observation space $S_c$ and a common action space $A_c$.
    \item A finite set of node types $T$.
    \item For each node type $t \in T$, a local attribute space $C(t)$, a local observation space $S(t)$ and a local action space $A(t)$.
\end{enumerate}

A universe describes a set of environments and a morphology describes one instance of environment. Given an environment $(S, A, \text{step})$, a morphology $m$ is given by

\begin{enumerate}
    \item common attributes $c \in C_c$;
    \item a graph $(V, E)$, together with a type $t(v) \in T$ and local
          attributes $c_v \in C(t(v))$ for each node $v \in V$.
    \item an observation decomposition
          $\mathsf{split} \colon S \to S_c \times \prod_{v \in V} S(t(v))$;
    \item an action recomposition
          $\mathsf{join} \colon A_c \times \prod_{v \in V} A(t(v)) \to A$.
\end{enumerate}

This representation operates directly on the structure of the environment and naturally supports a large set of model architectures such as graph neural networks, heterogeneous graph Transformers \citep{hao2024} and type-heterogeneous encoder–decoder models \citep{ma2020}. The separation between the morphological universe and the morphology also delineates clearly what a multi-morphology algorithm must be generic over, and when: on the morphological universe at algorithm instantiation time and on morphologies at specialization time. 

\btob{} defines a single morphological universe and each building in the dataset provides a morphology in addition to the standard Gymnasium environment. In the \btob{} universe, quantities that exist once per building (outdoor weather, calendar information, and energy consumption) are modeled as dedicated observation-only node types rather than placed in the common spaces; the common attribute space is reserved for static building-level information such as the construction year. The remaining node types comprise thermal zones and the various types of HVAC equipment. The type ''thermal zone'' defines local attributes $C(t)$ that contain, for instance, the window-to-wall ratio.

\section{Benchmark Problems}
\label{sec:benchmark}

\begin{figure}
    \centering
    \includegraphics[width=0.7\linewidth]{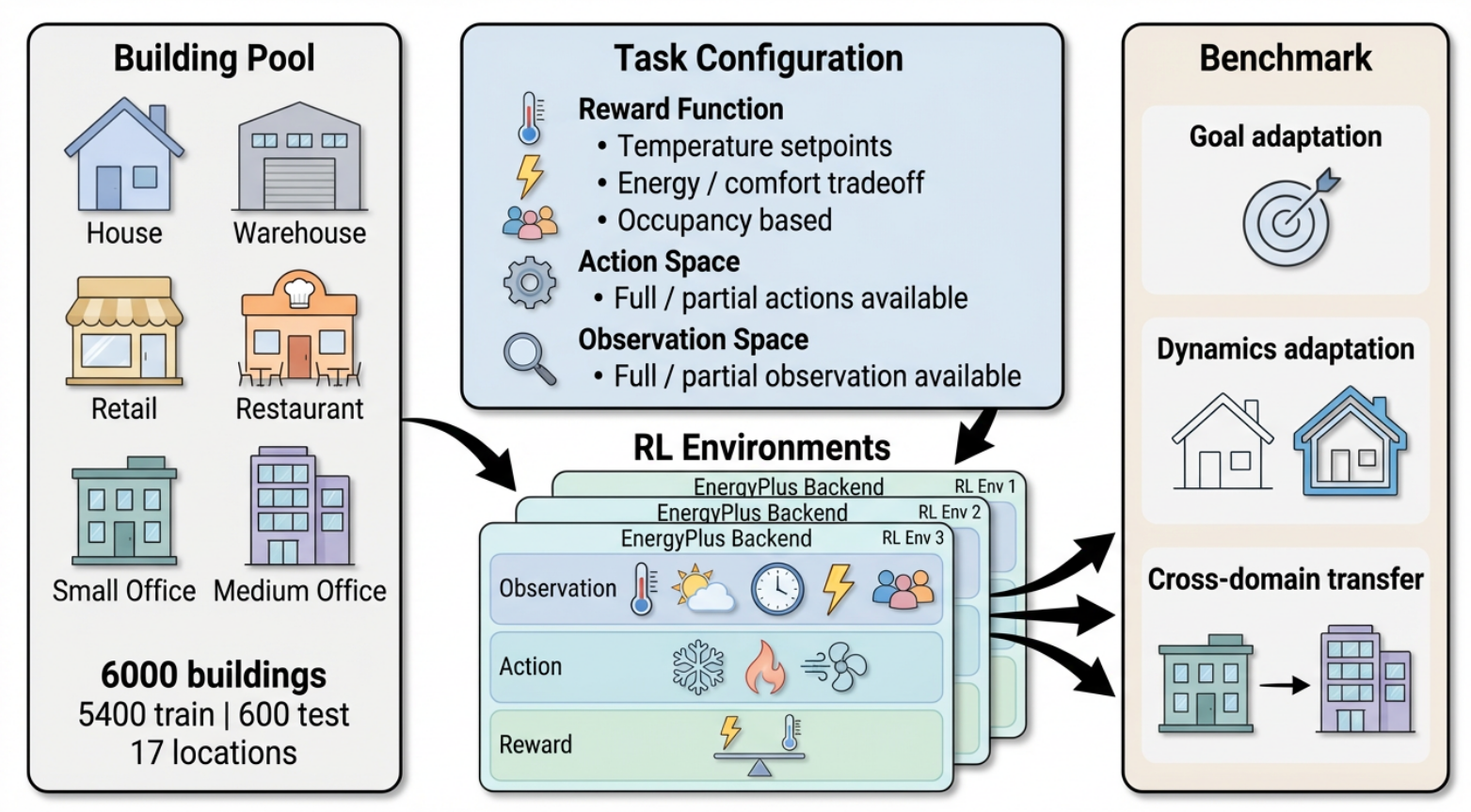}
    \caption{Overview of the benchmark design. A large building pool is created using our building generator. Different benchmark tasks evaluate transfer across goals, dynamics, and domains}
    \label{fig:b2b_benchmark}
\end{figure}

To define a reproducible experimental framework, we construct a diverse collection of environments by sampling buildings from the two generators described in \Cref{sec:B2B_suite}. For each building type, we generate 1000 environments, resulting in a total dataset of 6000 environments, split into a training and a test set. The resulting environment suite spans a wide range of building dynamics, HVAC configurations, and control interfaces. Additional details on the environments generation are given in \Cref{apx:building_dataset_generation}.

Using these environments, we define four benchmark settings that isolate different sources of variation in control systems: goal adaptation, dynamics variation, action-space shifts, and cross-domain generalization. These benchmarks are easily accessible through the API interface of \btob{}.

\begin{itemize}

\item[$\bullet$] \textbf{Goal adaptation.}  
The training and test environments correspond to the same building and control interface, but the reward function changes. Six tasks are defined from the Cartesian product of the following predefined reward settings:

\begin{itemize}
    \item[$\bullet$] \textbf{Energy weight} $w_E \in \{0.0, 0.5\}$, corresponding to the absence or presence of an energy consumption penalty in the reward function. The choice of weights is discussed in \Cref{apdx:reward_normalization}.
    
    \item[$\bullet$] \textbf{Zone temperature setpoints} (1) constant setpoints ($21^\circ C$); (2) occupancy-based setpoints ($21^\circ C$ when occupied, $18^\circ C$ or $26^\circ C$ otherwise, depending on the season); (3) random setpoints\footnote{Random setpoints do not represent typical occupant behaviour. However, it increases the difficulty of the task as it de-correlates setpoints from the time of day and day of the week.} .

\end{itemize}

\item[$\bullet$] \textbf{Dynamics adaptation - \Cref{tab:dynamics_adaptation}}. The reward function and action space remain fixed, while the building dynamics vary. Buildings differ in climate zone, size, and envelope but share the same archetype. The different settings have been chosen to vary the complexity of the dynamics : houses have a single thermal zone, small offices have multiple zones thermally coupled but with independent HVAC systems, and medium offices have multiple zones thermally coupled with centralized HVAC systems in which some actions impact simultaneously multiple zones.

\begin{table}[]
\caption{Dynamics Adaptation Settings.}
\label{tab:dynamics_adaptation}
\centering
\small
\begin{tabular}{lccc}
\toprule
\textbf{Setting} & \textbf{Building archetype} & \textbf{Variation} & \textbf{Action dim.} \\
\midrule
1   & Single-zone house & Climate, building envelope parameters & 2 \\
2 & Small office      & Climate, building envelope parameters & 10 \\
3   & Medium office     & Climate, building envelope parameters & 36 \\
\bottomrule
\end{tabular}
\end{table}

\item[$\bullet$] \textbf{Action-space transfer - \Cref{tab:action_space_transfer}}.  The building dynamics and reward function remain fixed, but the set of controllable actuators changes between training and testing.

\begin{table}
\caption{Action Space Transfer Settings.}
\label{tab:action_space_transfer}
\centering
\small
\begin{tabular}{lccc}
\toprule
\textbf{System type} & \textbf{Training control} & \textbf{Test control} & \textbf{Action dim.} \\
\midrule
Unitary system & Air flow rate & Air flow + SAT & $5 \rightarrow 10$ \\
Central system & VAV boxes only & VAV boxes + central SAT & $33 \rightarrow 36$  \\
Unitary system &  Air flow + SAT & Air flow rate & $10 \rightarrow 5$  \\
Central system & VAV boxes + central SAT & VAV boxes only & $36 \rightarrow 33$ \\
\bottomrule
\end{tabular}
\end{table}

\item[$\bullet$] \textbf{Cross-domain generalization - \Cref{tab:cross-domain_generalization}}. Agents are trained and tested on different environments with different observation and action spaces. The predefined settings have been chosen to cover the different types of buildings and HVAC systems.  

\begin{table}
\caption{Cross-domain generalization settings}
\label{tab:cross-domain_generalization}
\centering
\small
\begin{tabular}{lccc}
\toprule
\textbf{Setting} & \textbf{Train building type} & \textbf{Test building type} & \textbf{Key difference} \\
\midrule
1   & Retail store & Small office  & Similar HVAC types \\
2 & Retail store & Warehouse     & Slightly different HVAC types \\
3   & Small office & Medium office & Different HVAC types \\
4   & n types      & m different types & Different domains \\
\bottomrule
\end{tabular}
\end{table}

\end{itemize}

\section{Baselines}
\label{sec:baselines}

To establish a baseline performance, we implement supervisory control logic inspired by the ASHRAE Guideline 36 high-performance sequences of operation. Two controllers are implemented: one for unitary systems and one for central air-loop systems. The control structures include zone-level PI loops that regulate airflow and Trim-and-Respond mechanisms that adjust the SAT. Additional implementation details are provided in \Cref{apdx:baseline_controllers}, and detailed control results are reported in \Cref{apdx:additional_experiments-reactive_control_logic}. Reactive control strategies are widely used in real buildings, and we therefore use them as the primary baseline. To facilitate evaluation, the benchmark includes a function \textit{compute\_normalized\_score} that divides the cumulative return over a given period of simulation by the return obtained by the reactive controller, providing an easily interpretable normalized score.

In addition, we train PPO agents on a subset of the test buildings for each building type. The subset contains 8 buildings, one per climate zone. For single-zone houses, which all belong to the same climate zone, 8 buildings are randomly sampled from the test set. Agents are trained on the four goal adaptation tasks defined in \Cref{sec:benchmark}. Detailed results and normalized scores are reported in \Cref{apdx:additional_experiments-reactive_control_logic}.

\section{Generalization Experiments}
\label{sec:limitation_of_current_algorithms}

\subsection{Dynamics Adaptation}

\begin{figure}
    \centering
    \begin{subfigure}[b]{0.46\linewidth}
        \centering
        \includegraphics[width=\linewidth]{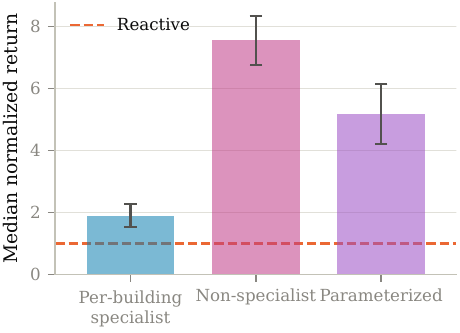}
        \caption{Median normalized return on the test buildings. The dashed line represents the reactive control baseline performance.}
        \label{fig:transfer_rew}
    \end{subfigure}
    \hfill
    \begin{subfigure}[b]{0.46\linewidth}
        \centering
        \includegraphics[width=\linewidth]{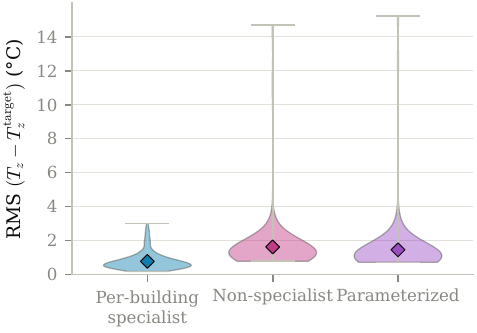}
        \caption{Violin plot of the RMS of temperature deviation from target temperature by model across test buildings. Diamonds represent the overall mean.}
        \label{fig:transfer_temp_deviation}
    \end{subfigure}
    \caption{Performances of three types of models for dynamics adaptation}
    \label{fig:transfer_side_by_side}
\end{figure}

In order to assess the performance of current methods on dynamics adaptation, we consider the single-zone houses (Setting 1, \Cref{tab:dynamics_adaptation}). We train three different classes of models to track a constant setpoint with no penalty on the energy (i.e. $w_E=0$) with PPO: (1) 100 per-building specialists that are trained solely on each building in the test set to establish an upper bound, (2) a non-specialist agent that resamples new buildings from the training set after every episode, and (3) a parameterized agent that resamples new buildings and has its observation space augmented with the building's floor area, year built, number of actuators and units. Implementation details are given in ~\Cref{apdx:dynamics_adaptation}.

\Cref{fig:transfer_rew} shows the median normalized return for each type of model (with per-building results only on the building that model was trained on). As expected, the per-building specialist achieves the lowest (best) score, with a substantial gap from the parameterized and non-specialist models. The gap between the non-specialist and parameterized models shows the benefit of conditioning on building parameters for dynamics adaptation. As shown in \Cref{fig:transfer_temp_deviation}, all models are able to control the temperature around the target. However, the non-specialist and parameterized models have a long tail distribution indicating a few cases among the 100 test buildings in which the transfer failed. On the other hand, per-building specialists performances are consistent throughout the test set.

\subsection{Cross domain transfer}

To demonstrate how \btob{} can be used to study cross-morphology control algorithms, we train a single policy simultaneously on four building types: retail store, fast food restaurant, small office and medium office. These types of buildings have heterogeneous action and observation spaces, making the environment incompatible with the standard multi-layer perceptron with fixed input and output sizes. We implement a variant of the Amorpheus~\citep{kurin2021} architecture that we train on top of the structured environment representation described in \Cref{sec:structured-env}.

The policy architecture consists of standard acausal transformer layers together with type-specific linear encoder and decoder blocks. At inference time, the observation is first split into node-level observations, which are encoded into a shared embedding space. The resulting embeddings are processed by the transformer, decoded into node-level actions, and finally concatenated to produce the global actuator command. Algorithm details are given in \Cref{apdx:cross_domain}.

\begin{figure}
    \centering
    \begin{subfigure}[b]{0.265\linewidth}
        \includegraphics[width=\linewidth]{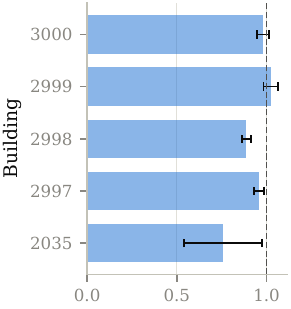}
        \caption{Retail}
    \end{subfigure}\hfill
    \begin{subfigure}[b]{0.235\linewidth}
        \includegraphics[width=\linewidth]{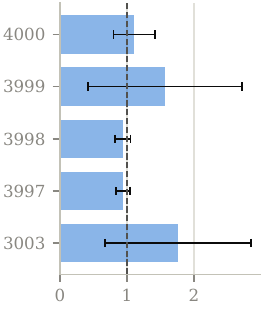}
        \caption{Restaurant}
    \end{subfigure}\hfill
    \begin{subfigure}[b]{0.235\linewidth}
        \includegraphics[width=\linewidth]{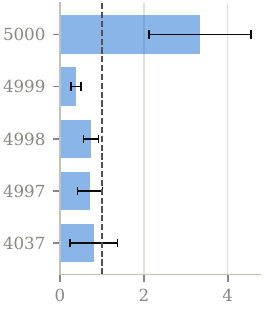}
        \caption{Medium Office}
    \end{subfigure}\hfill
    \begin{subfigure}[b]{0.235\linewidth}
        \includegraphics[width=\linewidth]{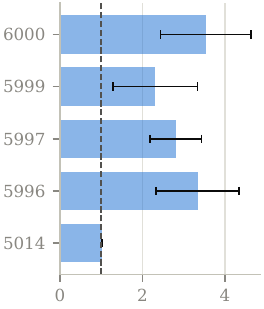}
        \caption{Small Office}
    \end{subfigure}

    \vspace{2pt}
    {\small Normalized return}

    \caption{Generalized policy test performance on 20 unseen buildings sampled
    from the test set. Return is normalized by the reactive baseline on the same
    episode window (dashed line at $1.0$); a value below $1$ beats the baseline.}
    \label{fig:multi-morphology}
\end{figure}

The model is trained to track a dynamic setpoint while ignoring energy costs (i.e. $w_E=0$). Training data are collected in parallel on four buildings (one of each type) during winter. Buildings are resampled every 10 PPO updates ensuring that the policy is exposed to the full diversity of buildings in the training set. The model is trained on 1M environment steps and tested on new buildings sampled from the test set. As shown in \Cref{fig:multi-morphology} the policy is able to adapt to new unseen buildings, and even outperform the baseline reactive controller on some of them.

\section{Conclusion}

We introduce \btob{}, a large-scale suite of environments and benchmark problems designed to advance research on generalization in RL. \btob{} provides a diverse set of environments with varying dynamics and heterogeneous observation and action spaces, making it well suited for studying multi-task learning and cross-domain transfer. The benchmark tasks cover several forms of adaptation, including goal adaptation and action-space shifts, and provide a reproducible framework for comparing algorithms.

Grounded in real-world applications, the control tasks focus on HVAC systems in buildings. Improving control in this domain has the potential to reduce energy consumption and contribute significantly to climate change mitigation.

\appendix

\subsubsection*{Acknowledgments}
\label{sec:ack}
This research was enabled in part by compute resources provided by Mila (mila.quebec).

\bibliography{main}
\bibliographystyle{rlj}

\beginSupplementaryMaterials

\section{Prior work on generalization in RL}
\label{apdx:prior_work_generalization_rl}

Generalization in RL has been widely studied. A common approach is Meta RL, where an agent is trained on a distribution of tasks so it can quickly adapt to new ones. Early methods include Model-Agnostic Meta-Learning (MAML) \citep{finn2017}, which adapts policies with a few gradient steps, and RL$^2$ \citep{duan2016}, which embeds the learning algorithm inside a recurrent network to share information across tasks via the RNN's hidden state. Later work introduced adaptation mechanisms such as attention-based meta-learners \citep{mishra2018}, and probabilistic context inference methods such as PEARL \citep{rakelly2019}, and Bayes-Adaptive RL (VariBAD) \citep{zintgraf2020}. 

A related line of work studies generalization through a multi-task formulation, where a single model is trained on multiple tasks simultaneously. Such an approach is challenging because gradients from different tasks interfere during learning. Methods such as PCGrad \citep{yu2020} mitigate gradient conflicts between tasks during training, while other approaches rely on multiple learner, such as policy distillation (Distral) \citep{teh2017} and mixture of orthogonal experts \citep{hendawy2024}. Large-scale systems such as IMPALA \citep{espeholt2018} and BRC \citep{nauman2025} demonstrated positive transfer across many tasks, while normalization techniques such as PopArt help address reward-scale imbalance across tasks\citep{hessel2018}.

Recent works also studies adaptation using sequence modeling and in-context reinforcement learning. In these approaches, models are trained on large datasets of trajectories and adapt to new tasks through conditioning on past experience in context. Seminal works in this direction include Decision Transformer,  \citep{chen2021}, Decision Pretrained Transformer \citep{lee2023} and Algorithm Distillation \citep{laskin2022}.

Most multi-task RL methods assume shared observation and action spaces across tasks. However, real-world problems often involve changes in morphology or environments. Methods such as DERL \citep{gupta2021}, MetaMorph \citep{gupta2022}, and AnyMorph \citep{trabucco2022} study control across agents with different morphologies. Several works study theoretically grounded transfer across domains. For instance successor representations and successor features enable reuse of value structure across related tasks \citep{Barreto2017,dayan1993}. Other approaches provide principled alignment across domains, for example through optimal-transport-based imitation learning \citep{fickinger2022}. 

\section{Building dataset generation}
\label{apx:building_dataset_generation}

We generate a climate-consistent building dataset using climate-dependent parameter ranges derived from ASHRAE 90.1-2022 Tables 5.5-1 through 5.5-8. Instead of sampling all seven envelope and geometry parameters independently over uniform global bounds, each building’s location is first mapped to its ASHRAE climate zone, and LHS unit samples are then remapped to zone-specific ranges defined relative to the prescriptive maxima for fenestration and framed walls. Window U-factors are sampled within $[0.8,\,1.3]\times$ the ASHRAE maximum for the given zone; SHGC is sampled between 0.15 and $(\text{SHGC}_{\max} + 0.10)$, clamped to $[0.10,\,0.80]$; and envelope conductivity and infiltration multipliers are upper-bounded as a function of climate stringency, ensuring tighter constructions in colder zones (ASHRAE 90.1-2022, \S5.4.3.1; Tables 5.5-1--5.5-8). Geometry scaling is narrowed to $[0.7,\,1.5]$ to avoid extreme HVAC autosizing artifacts, while orientation remains uniformly sampled in $[0^\circ,\,360^\circ]$. LHS sampling is preserved within each climate zone by first generating unit hypercube samples and then mapping them through zone-specific bounds. Finally, balanced train/test splits are produced with a dedicated script that performs stratified sampling by location (90/10), ensuring proportional climate representation across splits.

\section{Environments details}
\label{apdx:environments_details}

\buildingtobuilding{} converts building simulation models into standardized RL environments. Starting from building descriptions encoded as EnergyPlus input files, \btob{} automatically constructs a controllable simulation environment compatible with the Gymnasium API \citep{towers2024}. The pipeline parses the building model, identifies thermal zones and HVAC systems, exposes relevant control actuators, and generates standardized observation variables. This process abstracts away simulator-specific details and produces environments with consistent interfaces for observations, actions, and rewards. As a result, RL agents can be evaluated across a diverse set of buildings without requiring any simulator-specific implementations.

\begin{table}
\centering
\caption{Comparison of building control environments}
\label{tab:hvac_envs}

%\small % or \footnotesize for even smaller
\setlength{\tabcolsep}{4pt} % default is 6pt
\begin{tabular}{@{}lcccc@{}}
\toprule
\textbf{Environment} & \textbf{Buildings} \\
\midrule
RL Testbed \citep{moriyama2018} & 1   \\
BOPTEST(-Gym) \citep{blum2021} & $\sim$12  \\
Energym \citep{scharnhorst2021} & 14   \\
Sinergym \citep{campoy2025} & $\sim$30  \\
Smart bldg. Control \citep{goldfeder2025}  & 11  \\
CityLearn \citep{nweye2024} & Variable  \\
\textbf{Building2Building (Ours)} & \textbf{Thousands}   \\
\bottomrule
\end{tabular}
\end{table}

The resulting Gymnasium environment interacts with the EnergyPlus simulation using callbacks at every step to fetch observations and pass actions. An EnergyPlus timestep takes about $0.1$ms, and multiple EnergyPlus simulations can be launched in parallel. Timestep durations for the different building types are given in \Cref{tab:step_time}.

Other HVAC control environments based on high-fidelity open-source models, such as Sinergym \citep{campoy2025}, provide only a limited number of buildings. This limitation arises from two main challenges. First, EnergyPlus simulations require detailed input building descriptions, which are rarely available at large scale. Second, interacting with a complex simulator such as EnergyPlus in a consistent way is difficult, especially when exposing low-level HVAC actuators for control. \btob{} addresses these challenges in two ways. First, it provides a generator of EnergyPlus building files that enables the creation of numerous building environments in a principled way, allowing sensors and actuators to be extracted automatically. Second, it includes a processing pipeline that automatically converts EnergyPlus models into Python RL environments following the Gymnasium interface, allowing direct use with existing RL algorithms.

\begin{table}[ht]
\centering
\caption{EnergyPlus simulation step time per building type, computed over 10000 steps}
\label{tab:step_time}
\begin{tabular}{l r r r r}
\toprule
Building Type & Mean (ms/step) & Std (ms/step) \\
\midrule
Office Small & 0.170 & 0.018 &  \\
Office Medium & 0.410 & 0.096 & \\
Retail Standalone & 0.153 & 0.021 & \\
Restaurant Fast Food & 0.059 & 0.006 \\
Warehouse & 0.209 & 0.044  \\
Single-Family House & 0.021 & 0.002 \\
\bottomrule
\end{tabular}
\end{table} 

\section{Baseline controllers}
\label{apdx:baseline_controllers}

\subsection{Reactive Control Logic for Unitary Systems}
\label{apdx:reactive_control_logic_unitray}

 For zones equipped with a unitary system, the action space is the supply fan air mass flow rate and the supply air temperature setpoint. The control logic follows supervisory principles of ASHRAE Guideline-36, adapted to a 5-minute timestep. The zone temperature is regulated using a PI controller that modulates the supply air flow rate. Let $T_z$ denote the zone air temperature and $T_{sp}$ the active temperature setpoint (cooling or heating). The control error is defined as

    \begin{equation}
    e(t) = T_z(t) - T_{sp}(t).
    \end{equation}
    
    The airflow command is
    
    \begin{equation}
    \dot m_{sa}(t) =
    \dot m_{\min}
    + \left[
    K_p e(t) +
    K_i \int_{0}^{t} e(\tau)\, d\tau
    \right]
    \left(\dot m_{\max} - \dot m_{\min}\right),
    \end{equation}
    
    subject to saturation:
    
    \begin{equation}
    \dot m_{sa} \in [\dot m_{\min}, \dot m_{\max}].
    \end{equation}
    
    The gains $K_p$ and $K_i$ are tuned for stability under a 5-minute timestep.
    When the zone temperature lies within the deadband, airflow is reduced to
    $\dot m_{\min}$.

    The supply air temperature setpoint is adjusted using a Trim-and-Respond logic based on the zone demand. For instance, in a cooling regime:

    \begin{equation}
    T_{sa,sp}^{k+1} =
    \begin{cases}
    T_{sa,sp}^{k} - \Delta_{\text{resp}}, & \text{if } e_c^k > \delta \\
    T_{sa,sp}^{k} + \Delta_{\text{trim}}, & \text{otherwise}
    \end{cases}
    \end{equation}

    subject to bounds:
    
    \begin{equation}
    T_{sa,sp} \in [T_{sa,\min},\, T_{sa,\max}].
    \end{equation}
    
    Here $\Delta_{\text{resp}}$ and $\Delta_{\text{trim}}$ represent the respond and trim
    increments, respectively.

\subsection{Reactive Control Logic for Air Loops}
\label{apdx:reactive_control_logic_air_loop}
For zones served by a central air loop, the action space consists of the central supply air temperature setpoint $T_{sa,sp}$ and the per-zone damper opening fraction $\alpha_i \in [0,1]$. Each zone $i$ is equipped with a thermostat that modulates the local reheat coil. The zone temperature is regulated through the damper position using a PI controller. Let $T_{z,i}$ denote the zone air temperature and $T_{sp,i}$ the active temperature setpoint. The control error is

\begin{equation}
e_i(t) = T_{z,i}(t) - T_{sp,i}(t).
\end{equation}

The commanded damper opening fraction is

\begin{equation}
\alpha_i(t) =
\alpha_{\min}
+
\left[
K_p e_i(t) +
K_i \int_0^t e_i(\tau)\, d\tau
\right]
(\alpha_{\max} - \alpha_{\min}),
\end{equation}

subject to

\begin{equation}
\alpha_i \in [\alpha_{\min}, \alpha_{\max}].
\end{equation}

When the zone temperature lies within the deadband, the damper position is
maintained at $\alpha_{\min}$.

Each zone thermostat activates the reheat coil when the zone temperature falls
below the heating setpoint. The reheat power $Q_{rh,i}$ is modulated
proportionally to the heating error

\begin{equation}
Q_{rh,i}(t) = K_{rh}\,[T_{sp,i}(t) - T_{z,i}(t)]_+ ,
\end{equation}

where $[x]_+ = \max(0,x)$.

The central supply air temperature setpoint is adjusted using a Trim-and-Respond
logic based on aggregate cooling demand:

\begin{equation}
T_{sa,sp}^{k+1} =
\begin{cases}
T_{sa,sp}^{k} - \Delta_{\text{resp}}, & \text{if } D_c^k > \delta \\
T_{sa,sp}^{k} + \Delta_{\text{trim}}, & \text{otherwise}
\end{cases}
\end{equation}

subject to

\begin{equation}
T_{sa,sp} \in [T_{sa,\min}, T_{sa,\max}].
\end{equation}

\subsection{PPO implementation}
\label{apdx:ppo_implementation}

We use Stable-Baselines3 \citep{stable-baselines3} for the PPO implementation. Rollouts are collected on 14 environments in parallel. The training is done over 5M environment steps. The hyperparameters used are summarized in \Cref{tab:ppo_hyperparameters}.

\begin{table}[h]
\centering
\small
\caption{Hyperparameters used for PPO training.}
\label{tab:ppo_hyperparameters}
\begin{tabular}{ll ll}
\toprule
\textbf{Parameter} & \textbf{Value} & \textbf{Parameter} & \textbf{Value} \\
\midrule
Policy type & MLP & Learning rate & $5\times10^{-5}$ \\
Batch size & 336 & $n\_steps$ & 672 \\
Target KL & 0.02 & $n\_epochs$ & 5 \\
$\gamma$ & 0.98 & $\lambda$ (GAE) & 0.95 \\
Clip range & 0.2 & Entropy coef. & 0.01 \\
Value function coef. & 0.5 & Max grad norm & 0.5 \\
Use SDE & False & SDE sample freq. & 96 \\
\midrule
\multicolumn{4}{l}{\textbf{Policy network architecture}} \\
\midrule
Actor layers & [256, 256] & Critic layers & [256, 256] \\
Activation & Tanh & Orthogonal init & True \\
Initial log std & -1.0 & & \\
\bottomrule
\end{tabular}
\end{table}

\section{Reward normalization}
\label{apdx:reward_normalization}

Recall the reward definition given by (\Cref{eq:reward_function}):

\begin{equation*}
r_t = - \frac{1}{Z \, \tau_T} \sum_{z=1}^Z 
 (T_z(t) - T^{target}_z(t))^2
- \frac{w_E}{\tau_E} \, E_{\text{HVAC}}(t).
\end{equation*}

A difference of $1^\circ C$ in zone temperature has the same meaning across buildings and is easily  interpretable, we thus fix $\tau_T = 1$. The energy consumption term however depends on the building type, the building size, the HVAC type and the climate zone. To normalize the scale of the reward, we fix $\tau_E$ as the mean per-step power consumption of the baseline reactive policy of a given building type and climate zone. $\tau_E$ is building and climate dependent, and allows controlling the scale of the energy penalty term in the sum. $E_{\text{HVAC}}(t) / \tau_E$ is dimensionless and $w_E$ has a physical interpretation : how much temperature deviation (in $^\circ C)$ is worth a unit of energy consumption (with respect to the baseline power consumption). 

\begin{figure}[htbp]
    \centering

    % Row 1
    \begin{subfigure}{0.4\textwidth}
        \centering
        \includegraphics[width=\linewidth]{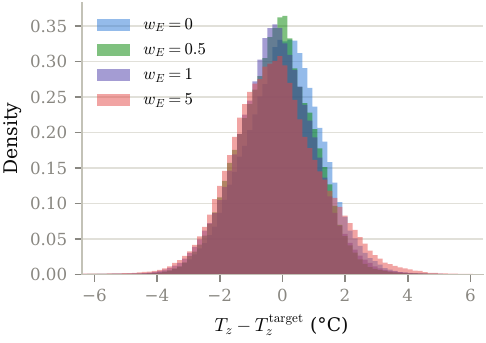}
        \caption{Medium Office - Constant setpoints}
        \label{fig:reward_temp_dist_medium_const}
    \end{subfigure}
    \hspace{1em}
    \begin{subfigure}{0.4\textwidth}
        \centering
        \includegraphics[width=\linewidth]{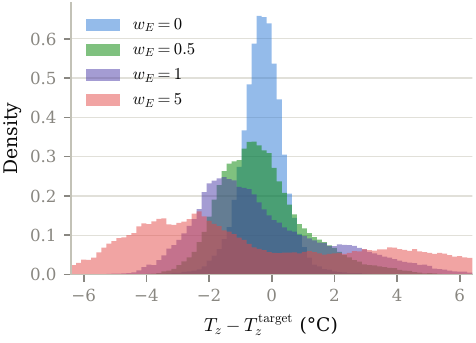}
        \caption{Small Office - Constant setpoints}
        \label{fig:reward_temp_dist_small_const}
    \end{subfigure}

    \vspace{0.5cm}

    % Row 2
    \begin{subfigure}{0.4\textwidth}
        \centering
        \includegraphics[width=\linewidth]{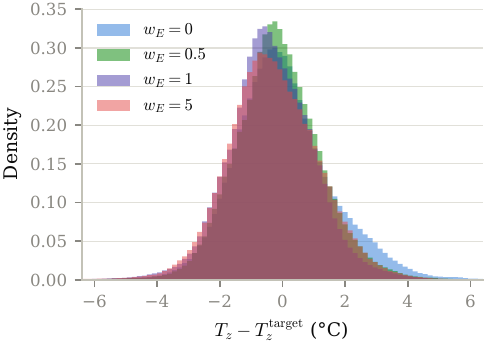}
        \caption{Medium Office - Occupancy based setpoints}
        \label{fig:reward_temp_dist_medium_occ}
    \end{subfigure}
    \hspace{1em}
    \begin{subfigure}{0.4\textwidth}
        \centering
        \includegraphics[width=\linewidth]{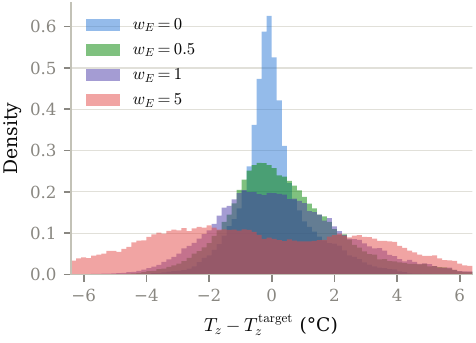}
        \caption{Small Office - Occupancy based setpoints}
        \label{fig:reward_temp_dist_small_occ}
    \end{subfigure}

    \caption{Temperature distribution for different energy penalty weights}
    \label{fig:reward_temp_dist}
\end{figure}

\begin{figure}[htbp]
    \centering

    % Row 1
    \begin{subfigure}{0.45\textwidth}
        \centering
        \includegraphics[width=\linewidth]{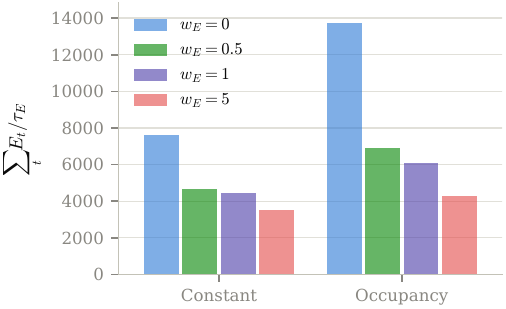}
        \caption{Medium Office}
        \label{fig:reward_energy_medium}
    \end{subfigure}
    \hspace{1em}
    \begin{subfigure}{0.45\textwidth}
        \centering
        \includegraphics[width=\linewidth]{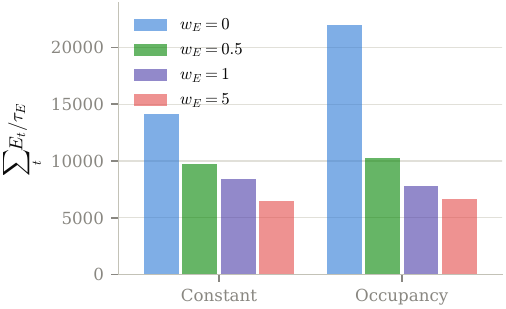}
        \caption{Small Office}
        \label{fig:reward_energy_small}
    \end{subfigure}

    \caption{Energy consumption for different energy penalty weights}
    \label{fig:reward_energy}
\end{figure}

We investigate the behaviour of learned PPO policies with respect to the choice of $w_E \in \{0; 0.5; 1.0; 5.0 \}$ on two building types, as shown in \Cref{fig:reward_temp_dist} and \Cref{fig:reward_energy}. As expected, the higher $w_E$ the lower the total energy consumption. However, despite the normalization, we observe different impacts of $w_E$ on the temperature control : in the Medium office, the mean temperature deviation distribution does not shift much, but in the Small office for high values of $w_E$ the learned policy is degenerate and focus only on limiting power consumption by turning off the HVAC system. This highlights that the same choice of weights does not necessarily result in the same objectives trade-off depending on the environment.

The pre-defined tasks feature two choices of rewards : temperature control only ($w_E = 0$) and with an energy consumption penalty ($w_E=0.5$). We kept $w_E=0.5$ as it yields no degenerate policies on every building and climate zones of the small test set in both winter and summer periods. 

\section{Additional experiments}
\label{apdx:additional_experiments}

\subsection{Baselines results}
\label{apdx:additional_experiments-reactive_control_logic}

In this section we present the results of the reactive controller and PPO agents on the test buildings. 

\begin{figure}
    \centering
    % Row 1
    \begin{subfigure}{0.32\textwidth}
        \centering
        \includegraphics[width=\linewidth]{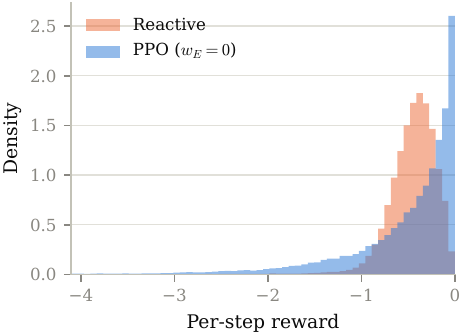}
        \caption{Small Office}
        \label{fig:const_sub1}
    \end{subfigure}
    \hfill
    \begin{subfigure}{0.32\textwidth}
        \centering
        \includegraphics[width=\linewidth]{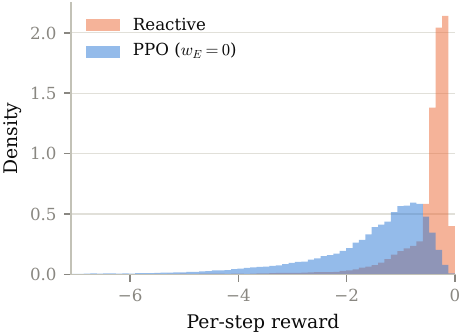}
        \caption{Medium Office}
        \label{fig:const_sub2}
    \end{subfigure}
    \hfill
    \begin{subfigure}{0.32\textwidth}
        \centering
        \includegraphics[width=\linewidth]{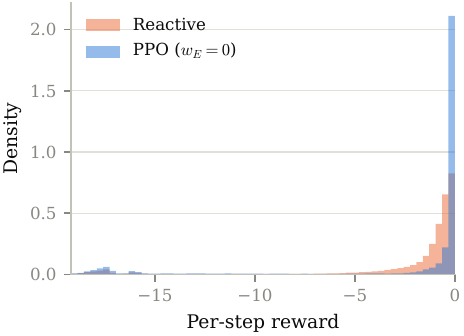}
        \caption{Fastfood}
        \label{fig:const_sub3}
    \end{subfigure}
    \vspace{0.5cm}
    % Row 2
    \begin{subfigure}{0.32\textwidth}
        \centering
        \includegraphics[width=\linewidth]{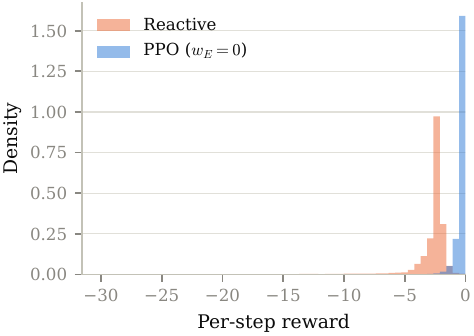}
        \caption{House}
        \label{fig:const_sub4}
    \end{subfigure}
    \hfill
    \begin{subfigure}{0.32\textwidth}
        \centering
        \includegraphics[width=\linewidth]{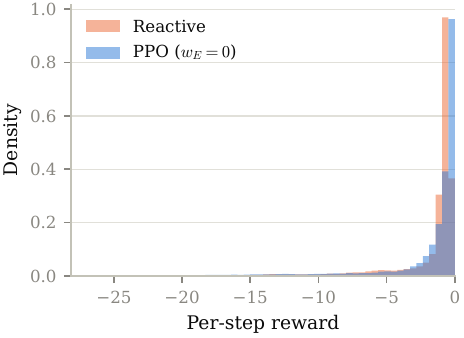}
        \caption{Warehouse}
        \label{fig:const_sub5}
    \end{subfigure}
    \hfill
    \begin{subfigure}{0.32\textwidth}
        \centering
        \includegraphics[width=\linewidth]{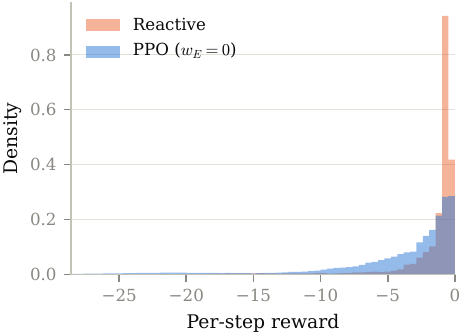}
        \caption{Retail}
        \label{fig:const_sub6}
    \end{subfigure}
    \caption{Per step reward distribution aggregated over the 8 buildings of the small test set for constant setpoints.}
    \label{fig:perstep_reward_constant}
\end{figure}

\begin{figure}
    \centering
    % Row 1
    \begin{subfigure}{0.32\textwidth}
        \centering
        \includegraphics[width=\linewidth]{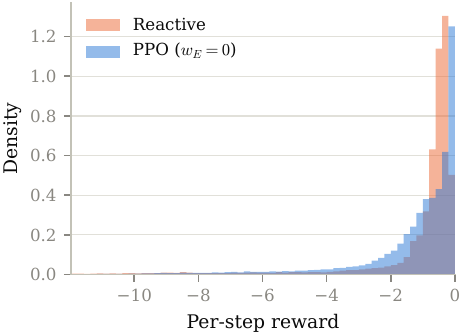}
        \caption{Small Office}
        \label{fig:occ_sub1}
    \end{subfigure}
    \hfill
    \begin{subfigure}{0.32\textwidth}
        \centering
        \includegraphics[width=\linewidth]{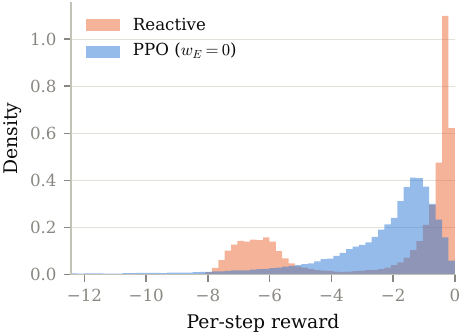}
        \caption{Medium Office}
        \label{fig:occ_sub2}
    \end{subfigure}
    \hfill
    \begin{subfigure}{0.32\textwidth}
        \centering
        \includegraphics[width=\linewidth]{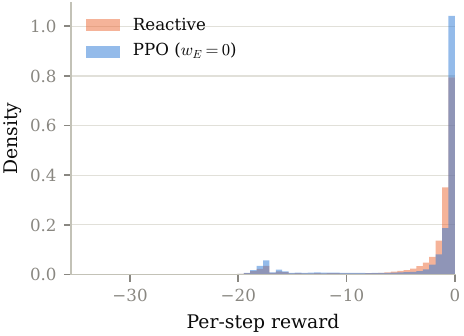}
        \caption{Fastfood}
        \label{fig:occ_sub3}
    \end{subfigure}
    \vspace{0.5cm}
    % Row 2
    \begin{subfigure}{0.32\textwidth}
        \centering
        \includegraphics[width=\linewidth]{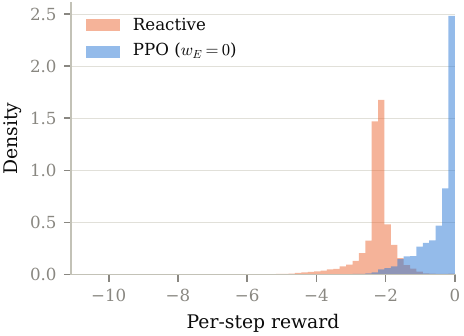}
        \caption{House}
        \label{fig:occ_sub4}
    \end{subfigure}
    \hfill
    \begin{subfigure}{0.32\textwidth}
        \centering
        \includegraphics[width=\linewidth]{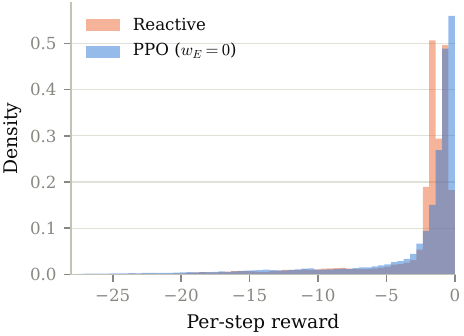}
        \caption{Warehouse}
        \label{fig:occ_sub5}
    \end{subfigure}
    \hfill
    \begin{subfigure}{0.32\textwidth}
        \centering
        \includegraphics[width=\linewidth]{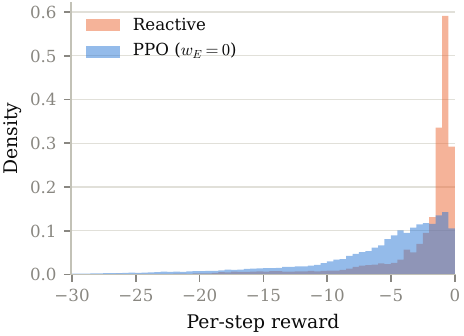}
        \caption{Retail}
        \label{fig:occ_sub6}
    \end{subfigure}
    \caption{Per step reward distribution aggregated over the 8 buildings of the small test set for occupancy based setpoints.}
    \label{fig:per_step_reward_occ}
\end{figure}

For the reactive controllers, hyperparameters of each controller were tuned on the test buildings using Bayesian optimization to optimize for $25^{th}$ percentile performance, with occupancy based setpoints and no energy penalty as the optimization objective. One set of hyperparameters is used per building type and climate zone. \Cref{fig:perstep_reward_constant} and \Cref{fig:per_step_reward_occ} present the performance of the reactive controllers. Reactive controllers are designed for temperature control only, we thus fix $w_E=0$ and the reward represents temperature deviation (in $^\circ C$). The distributions' modes are around $0$, indicating a good reward control. We explain the tail of the distribution by the fixed set of hyperparameters. Indeed, some buildings such as small buildings with low thermal inertia are more sensitive to specific choices of hyperparameter. 

PPO agents are trained specifically on each building of the small test set, across different tasks and seasons, using the hyperparameters of \Cref{apdx:ppo_implementation}. \Cref{tab:ppo_scores_ftau} shows the results of the  \Cref{fig:perstep_reward_constant} and \Cref{fig:per_step_reward_occ} compares the per-step reward distribution of the PPO agents to the reactive control baseline.

\begin{table}
  \centering
  \footnotesize
  \caption{Performances of PPO agents on the small test set for different tasks and seasons. The scores are normalized with respect to the reactive baseline performance. The \textit{Beats} columns indicates the number of test buildings in which PPO outperforms the baseline.}
  \label{tab:ppo_scores_ftau}
  \begin{tabular}{@{}l l c rrr c @{\hspace{10pt}} rrr c@{}}
    \toprule
    & & & \multicolumn{4}{c}{Winter} & \multicolumn{4}{c}{Summer} \\
    \cmidrule(lr){4-7} \cmidrule(l){8-11}
    Building type & Setpoint & $w_E$ & Median & Min & Max & Beats & Median & Min & Max & Beats \\
    \midrule
    \multirow{4}{*}{Office Small} & Constant & $0$ & 0.95 & 0.54 & 2.78 & 4/8 & 0.47 & 0.19 & 0.53 & 8/8 \\
     &  & $0.5$ & 2.09 & 1.24 & 4.94 & 0/8 & 1.05 & 0.57 & 2.07 & 2/8 \\
    \cmidrule(l){2-11}
     & Occupancy & $0$ & 1.19 & 0.62 & 1.52 & 2/8 & 1.35 & 0.37 & 1.89 & 1/8 \\
     &  & $0.5$ & 1.29 & 0.76 & 10.86 & 1/8 & 1.89 & 1.51 & 3.36 & 0/8 \\
    \midrule
    \multirow{4}{*}{Office Medium} & Constant & $0$ & 2.80 & 1.49 & 9.07 & 0/8 & 1.65 & 0.60 & 4.09 & 3/8 \\
     &  & $0.5$ & 1.58 & 0.93 & 2.35 & 1/8 & 0.95 & 0.59 & 2.03 & 4/8 \\
    \cmidrule(l){2-11}
     & Occupancy & $0$ & 1.04 & 0.57 & 1.32 & 4/8 & 0.31 & 0.25 & 0.39 & 8/8 \\
     &  & $0.5$ & 0.72 & 0.52 & 0.89 & 8/8 & 0.30 & 0.25 & 0.39 & 8/8 \\
    \midrule
    \multirow{4}{*}{Retail Standalone} & Constant & $0$ & 2.42 & 1.42 & 3.79 & 0/8 & 2.62 & 1.11 & 9.08 & 0/8 \\
     &  & $0.5$ & 1.97 & 1.55 & 7.87 & 0/8 & 2.98 & 1.13 & 10.10 & 0/8 \\
    \cmidrule(l){2-11}
     & Occupancy & $0$ & 2.28 & 1.54 & 5.01 & 0/8 & 2.29 & 0.62 & 3.19 & 1/8 \\
     &  & $0.5$ & 2.43 & 1.26 & 3.94 & 0/8 & 2.32 & 0.87 & 3.87 & 1/8 \\
    \midrule
    \multirow{4}{*}{Restaurant (Fast Food)} & Constant & $0$ & 0.26 & 0.19 & 1.54 & 6/8 & 1.12 & 0.93 & 1.26 & 2/8 \\
     &  & $0.5$ & 0.80 & 0.55 & 1.39 & 6/8 & 1.11 & 0.93 & 1.24 & 2/8 \\
    \cmidrule(l){2-11}
     & Occupancy & $0$ & 0.93 & 0.38 & 1.59 & 4/8 & 1.27 & 0.96 & 1.59 & 2/8 \\
     &  & $0.5$ & 0.94 & 0.72 & 1.49 & 4/8 & 1.22 & 0.96 & 1.47 & 2/8 \\
    \midrule
    \multirow{4}{*}{Warehouse} & Constant & $0$ & 1.03 & 0.44 & 1.54 & 3/8 & 2.32 & 1.10 & 4.18 & 0/8 \\
     &  & $0.5$ & 1.34 & 0.78 & 1.66 & 3/8 & 2.56 & 1.20 & 4.00 & 0/8 \\
    \cmidrule(l){2-11}
     & Occupancy & $0$ & 0.92 & 0.59 & 1.69 & 5/8 & 1.88 & 0.97 & 4.38 & 1/8 \\
     &  & $0.5$ & 1.01 & 0.71 & 1.73 & 4/8 & 2.04 & 0.88 & 3.97 & 1/8 \\
    \midrule
    \multirow{4}{*}{Single-Family House} & Constant & $0$ & 0.09 & 0.03 & 0.18 & 8/8 & 3.23 & 1.87 & 4.71 & 0/8 \\
     &  & $0.5$ & 0.26 & 0.06 & 0.44 & 8/8 & 3.63 & 2.70 & 4.72 & 0/8 \\
    \cmidrule(l){2-11}
     & Occupancy & $0$ & 0.18 & 0.11 & 0.31 & 8/8 & 1.02 & 0.60 & 1.86 & 4/8 \\
     &  & $0.5$ & 0.29 & 0.16 & 0.56 & 8/8 & 2.48 & 0.91 & 3.52 & 1/8 \\
    \bottomrule
  \end{tabular}
\end{table}

\subsection{Dynamics adaptation}
\label{apdx:dynamics_adaptation}

All buildings in this setting have a two-dimensional action space. Their observation spaces are padded to 20 dimensions, as the number of unconditioned zones (and hence of observed zone temperatures) varies across buildings. Episodes span 90 days of the winter period, i.e., 25,920 timesteps at a 5-minute resolution. All models are trained with PPO for 4 million environment steps, using policy and value networks with two hidden layers of 256 neurons and tanh activations. The per-building specialists (1) are each trained on a single test building. The non-specialist (2) and parameterized (3) models are trained on 16 environments running in parallel, each sampling a new building from the training set at every episode reset; both are trained with 9 random seeds. The parameterized model augments the observation with five normalized building parameters: floor area, construction year, number of warm-up phases, number of actuators, and number of units. In \Cref{fig:transfer_rew}, bars report the median normalized return over the 100 test buildings; error bars show the standard error across the 9 per-seed medians for the multi-building models, and a bootstrap standard error across buildings for the specialists.

\subsection{Cross-domain transfer}
\label{apdx:cross_domain}

Let $\mathrm{Emb} = \mathbb{R}^n$ denote the embedding space for a fixed embedding dimension $n$. The learnable components of the policy are (i) type-specific linear encoders $\{e_t : S(t) \to \mathrm{Emb}\}_{t \in T}$, (ii) a multi-layer transformer $\mathrm{trans} : \mathrm{Emb}^{|V|} \to \mathrm{Emb}^{|V|}$, and (iii) type-specific linear decoders $\{d_t : \mathrm{Emb} \to A(t)\}$ for the node types $t$.

Policy evaluation proceeds as follows:
\vspace{0.5em}

\begin{algorithmic}[1]
\Function{Policy}{$s_\text{raw}$}
    \State $s_v \gets \mathsf{split}(s_\text{raw})_v \quad \forall v$
    \State $i_v \gets e_{t(v)}(s_v) \quad \forall v$
    
    \State $o \gets \mathsf{trans}(i)$
    \State $a_v \gets d_{t(v)}(o_v)  \quad \forall v $
    \State \Return $\mathsf{join}(a)$
\EndFunction
\end{algorithmic}

\end{document}